\pdfoutput=1

\documentclass[11pt]{article}

\usepackage[]{emnlp2021}

\usepackage{times}
\usepackage{latexsym}

\usepackage[T1]{fontenc}

\usepackage[utf8]{inputenc}

\usepackage{microtype}
\usepackage{ifthen}
\usepackage{color}
\usepackage[utf8]{inputenc}
\usepackage{url}
\usepackage{multicol}
\usepackage{tikz}
\usepackage{booktabs}
\usepackage{csquotes}
\usepackage{soul}
\usepackage{float}
\usetikzlibrary{fit}
\usetikzlibrary{patterns}

\usepackage{enumitem}
\usepackage{tabularx}

\usepackage{booktabs}
\usepackage{amssymb}
\usepackage{xspace}

\usepackage{bbm}
\usepackage{amsmath}
\usepackage{url}
\usepackage{comment}
\usepackage{subcaption}

\usepackage{url}
\usepackage{enumitem}
\usepackage{booktabs}
\usepackage{multirow}
\usepackage{graphicx}
\usepackage{array}
\usepackage{subcaption}
\usepackage{amsthm,amssymb,amsopn,bm}
\usepackage{color,soul}
\usepackage{verbatim}
\usepackage{caption}
\usepackage{xcolor,colortbl}
\definecolor{green}{rgb}{0.1,0.1,0.1}
\usepackage{tabulary}
\usepackage{footnote}
\usepackage{blindtext}
\usepackage{algorithm}
\usepackage[noend]{algpseudocode}
\usepackage{stmaryrd}
\usepackage{CJKutf8}
\usepackage[OT2,T1]{fontenc}
\usepackage[russian, english]{babel}
\usepackage{multirow}
\makesavenoteenv{tabular}
\makesavenoteenv{algorithm}
\usepackage{tabularx,booktabs}
\newcolumntype{Y}{>{\centering\arraybackslash}X}

\setlist{leftmargin=8mm}

\usepackage{enumitem}

\newcommand{\boldcy}[1]{{\textbf{\{\color{cyan}{#1}}\}}}
\usepackage{tikz}

\definecolor{gitred}{HTML}{FDB8C0}
\definecolor{gitgreen}{HTML}{006400}
\definecolor{chocolate}{HTML}{D2691E}
\definecolor{maroon}{HTML}{800000}
\definecolor{indigo}{HTML}{4B0082}
\definecolor{green}{HTML}{008000}
\definecolor{orange}{HTML}{fc8d62}
\definecolor{purple}{HTML}{8da0cb}

\newcommand{\single}{$\mathbf{X}_{s} $ }
\newcommand{\unlabel}{$\mathbf{X}_{u} $ }
\newcommand{\multi}{$\mathbf{X}_{m} $ }
\newcommand{\singlen}{$\mathbf{X}_{s}$}

\newcommand{\multin}{$\mathbf{X}_{m}$}
\newcommand{\mxone}[1]{MixUp ($\mathbf{X}_{#1}$)}
\newcommand{\mxthree}[3]{MixUp ($\mathbf{X}_{#1}$, $\mathbf{X}_{#2}$, $\mathbf{X}_{#3}$)}
\newcommand{\mxtwo}[2]{MixUp ($\mathbf{X}_{#1}$, $\mathbf{X}_{#2}$)}

\title{
Learning with Different Amounts of Annotation: \\
From Zero to Many Labels

}

\author{Shujian Zhang  \qquad Chengyue Gong \qquad  Eunsol Choi \\
The University of Texas at Austin \\
\texttt{szhang19@utexas.edu, \{cygong,eunsol\}@cs.utexas.edu}
}

\begin{document}
\maketitle

\begin{abstract}
Training NLP systems typically assumes access to annotated data that has a single human label per example. Given imperfect labeling from annotators and inherent ambiguity of language, we hypothesize that single label is not sufficient to learn the spectrum of language interpretation. We explore new annotation distribution schemes, assigning multiple labels per example for a small subset of training examples. Introducing such multi label examples at the cost of annotating fewer examples brings clear gains on natural language inference task and entity typing task, even when we simply first train with a single label data and then fine tune with multi label examples. Extending a MixUp data augmentation framework, we propose a learning algorithm that can learn from training examples with different amount of annotation (with zero, one, or multiple labels). 
This algorithm efficiently combines signals from uneven training data and brings additional gains in low annotation budget and cross domain settings. Together, our method achieves consistent gains in two tasks, suggesting distributing labels unevenly among training examples can be beneficial for many NLP tasks.\footnote{Code and data split is available at \url{https://github.com/szhang42/Uneven_training_data}.}
\end{abstract}

\section{Introduction}\label{sec:intro}
Crowdsourcing annotations~\cite{Rajpurkar2016SQuAD10,Bowman2015ALA} has become a common practice for developing natural language processing benchmark datasets. Even after thorough quality control, it is often infeasible to reach complete annotator agreement, as annotators make mistakes~\cite{Freitag2021ExpertsEA} and ambiguity is a key feature of human communication~\cite{Asher2005LogicsOC}. 
Rich prior works~\cite{Passonneau2012MultiplicityAW,Pavlick2019InherentDI,nie2020can,Min2020AmbigQAAA,ferracane-etal-2021-answer} show that disagreement among annotators is not an annotation artifact but rather core linguistic phenomena.

\begin{figure}[t]
\centering
\includegraphics[width=7.8cm]{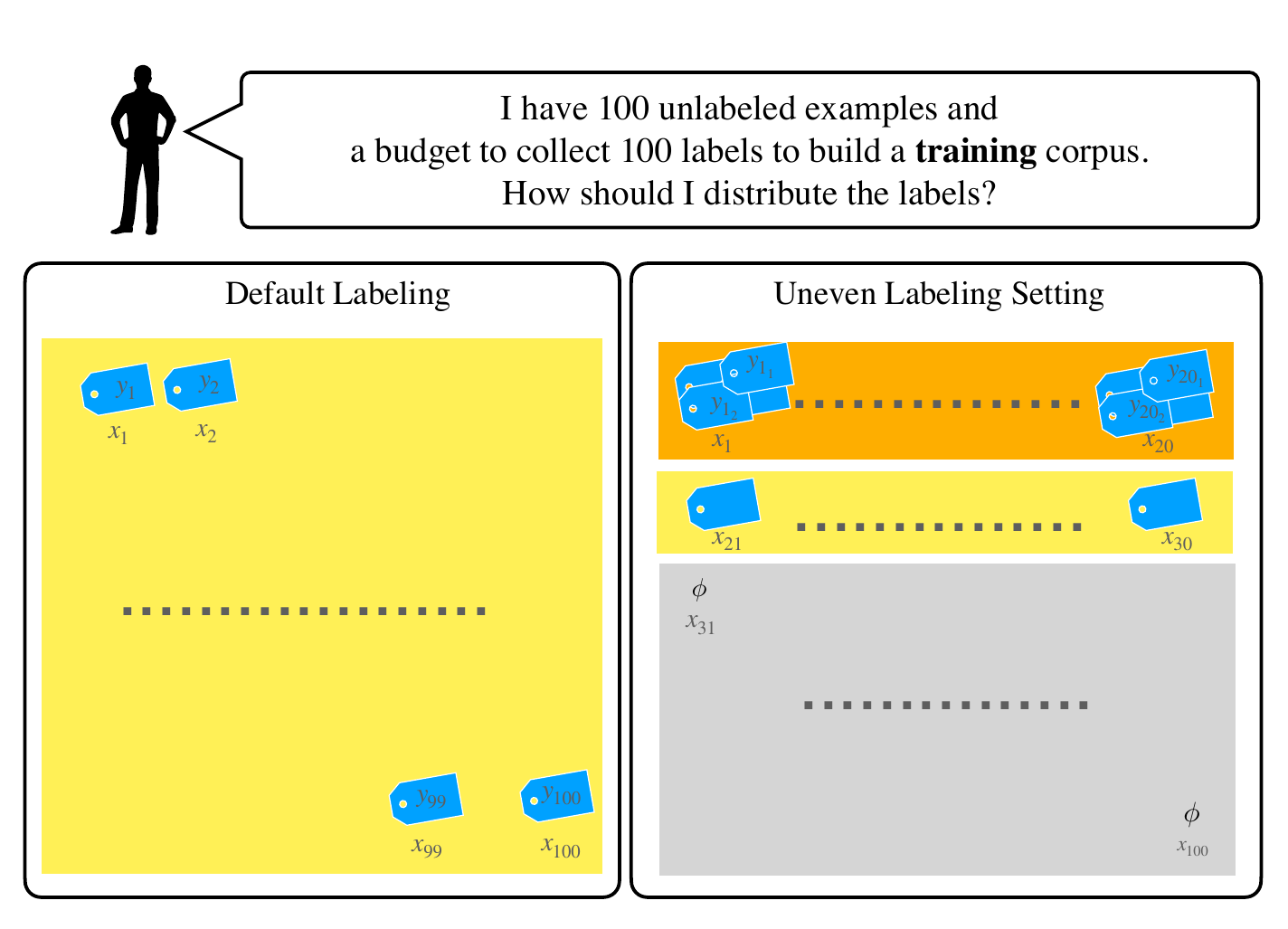} 
\caption{Re-thinking how to distribute annotation budget. Each blue tag represents a human annotation for the corresponding $x$. Examples in the orange shaded area are assigned many labels (multi label data), examples in the yellow shaded area are assigned a single label (single label data), and examples in grey shaded area are not assigned any labels. Models trained on a combination of multi label, single label and unlabeled data outperform models trained on single label data on both label accuracy and label distribution metrics for entailment and entity typing task. 
} 
\label{fig:intro}
\end{figure}

Despite observing such inherent ambiguity, most work have not embraced ambiguity into the training procedure. Most existing datasets~\cite{Wang2019SuperGLUEAS,Rajpurkar2016SQuAD10} present a single label per each training example while collecting multiple labels for examples in the evaluation set, with a few notable exceptions on subjective tasks~\cite{Passonneau2012MultiplicityAW,ferracane-etal-2021-answer}. We challenge this paradigm and re-distribute annotation budget unevenly among training examples, generating small amount of training examples with multiple labels. 
Without changing mainstream model architectures~\cite{transformer}, we change the annotation budget allocation. Figure~\ref{fig:intro} visualizes the standard scheme to our new label distribution scheme. 

Under our uneven label distribution scheme, models are given a mixture of single label, multi label and unlabeled examples as a training corpus. How should we combine learning signals from distinct types of training examples? We explore simply combining and shuffling examples, upsampling multi label examples, and curriculum learning. Then, we introduce an algorithm based on recent data augmentation MixUp~\cite{zhang2018mixup} which generates virtual training examples by interpolating between different training examples.

We present a retrospective study~\cite{Liu2021CanSA}
with datasets from prior work \cite{nie2020can,choi2018ultra}. We first evaluate our approach on densely annotated NLI datasets, where human disagreement is prevalent~\cite{Pavlick2019InherentDI}. We report majority label accuracy and distribution metrics (e.g., KL divergence to measures models' ability to estimate human label distribution). Our experiment on a multi label task -- fine-grained entity typing~\cite{choi2018ultra} -- exhibits similar trend that acquiring multiple labels for a single example is more effective than labeling as many examples as possible. 

Lastly, we present an in-depth study comparing models trained with multi label data and models trained with single label data. Training with single label examples leads the low entropy label distribution and unable to capture human disagreements. While calibration techniques such as smoothing distribution~\cite{Guo2017OnCO} can alleviate over confidence of model prediction and improves distributional metrics, it erroneously introduces uncertainty even for unambiguous examples. Our study suggests that introducing uneven label distribution scheme, paired with a learning architecture that combines three different types of training examples, can provide an efficient and effective solution. 


\section{Data Configuration}
We first describe our training data configuration and then discuss our learning algorithms. We notate the input feature vector as $x$ and output label distribution as $y$. We have three types of training example: unlabeled data set $\mathbf{X}_{u} = \{x_u^{1}, x_u^2 \ldots, x_u^{u_n})\}$, where $u_n$ is the total number of unlabeled examples, single label data set $\mathbf{X}_{s} = \{(x_s^{1}, y_s^{1}), (x_s^{2}, y_s^{2}) \ldots, (x_s^{s_n}, y_s^{s_n})\}$ where $s_n$ is the total number of single label examples, and multi label data set
\begin{align*}
    \mathbf{X}_{m} = & \{(x_m^{1}, (y_{m1}^{1}, y_{m_2}^{1} \ldots y_{m_k}^{1})) \ldots, \\
    &(x_m^{m_n}, (y_{m_1}^{m_n},  y_{m_2}^{m_n} \ldots y_{m_k}^{m_n}))\},
\end{align*}  where $m_n$ is the total number of multi label examples and $k$ is the number of annotations per example. For multi label examples, we will aggregate multiple annotations to generate $y_m^*$. Unlike $y_s$, which is a one-hot vector, $y_m^*$ will now be a distribution over labels (for label distribution estimation problem, averaging $(y_{m_1}^{i}, y_{m_2}^{i} \ldots y_{m_k}^{i})$, and for label prediction problem, taking $\arg\max_k (y_{m_1}^{i}, y_{m_2}^{i} \ldots y_{m_k}^{i})$).

The annotation cost for generating training datasets can be described as the function of two factors~\cite{Sheng2008GetAL}: the number of examples and the number of labels. Both can have impacts on the model performance and are highly associated with the annotation cost. In most existing studies~\cite{Wang2019SuperGLUEAS}, the training data is a set of annotated example with single label, $\mathbf{X}_{s}$. Supervised learning assumes an access to $\mathbf{X}_{s}$, and unsupervised learning assumes additional unlabeled examples $\mathbf{X}_{u}$, and semi-supervised learning assumes a mixture of $\mathbf{X}_{u}$ and $\mathbf{X}_{s}$. Here, we focus on annotation distribution over examples and make a simplifying assumption that annotation cost scales linearly to the number of labels.
\begin{table*}
\footnotesize
\begin{center}
\resizebox{2.1\columnwidth}{!}{\begin{tabular}{p{180pt}p{120pt}p{55pt}p{55pt}}
\toprule
\textbf{Premise} & \textbf{Hypothesis} & \textbf{Old Labels} & \textbf{New Labels}\\\midrule
A woman in a tan top and jeans is sitting on a
bench wearing headphones. & A woman is listening to music. & E E N N E & N (93) E (7)
\\ 
\bottomrule \vspace{+0.2mm}
\end{tabular}}
\resizebox{2.1\columnwidth}{!}{\begin{tabular}{p{300pt}p{120pt}} 
\toprule
\textbf{Sentence with Target Entity} & \textbf{Entity Type Labels}\\ \midrule
During the Inca Empire, \boldcy{the Inti Raymi} was the most important of four ceremonies celebrated in Cusco. & event, festival, {ritual, custom, ceremony, party, celebration}\\ 
\bottomrule
\end{tabular}}
\end{center}
\caption{Examples of ChaosSNLI and Ultra-fine Entity Typing dataset. In NLI task, each label corresponds to one annotator's judgement (entailment (E) / neutral (N) / contradiction (C)). In fine-grained entity typing, the entity mention is in blue with the curly brackets. Each positive type label is treated a single label. }
\label{tab:each_example}
\end{table*}

\begin{table*}
\centering
\footnotesize
 \begin{tabular}{l|l|c|c|c|c|c}
 \toprule
 {Task} & {Data Setup}&   \# {Single} & \# {Multi}  & \# {Unlabel}  & {Total \# Labels} & {Total \# Examples}\\ \midrule
 & Original & 549k / 392k & 0& 0 & 549k / 392k& 549k / 392k  \\ 
 & \single &150k & 0 & 0 &  150k * 1 = 150k& 150k\\
Chaos  & \single + \multi &  145k & 0.5k &  0& 145k * 1 + 0.5k * 10  = 150k & 145.5k \\
S / MNLI& \single + \unlabel & 150k  &0 & 549k-150k & 150k * 1  = 150k & 549k \\
  &  \single + \multi + \unlabel 
 & 145k & 0.5k & 549k-145.5k&  145k * 1 + 0.5k * 10  = 150k& 549k \\
 \midrule
 & Original & 151 & 1768 & 0 & 10.3k & 1919\\ 
& \single & 500 & 0 & 0  &  500 * 1 = 500 & 500   \\
UFET& \single + \multi& 100 & 200  &  0& 100 * 1 + 200 * 2 = 500  & 300  \\ 
 & \single + \unlabel  &  500 & 0 & 1919 - 500 & 500 * 1 = 500  & 1919\\
& \single + \multi + \unlabel & 100 & 200   &1919 - 300 &   100 * 1 + 200 * 2  = 500 & 1919 \\
 \bottomrule
  \end{tabular}
 \caption{Training data configurations. Each configuration is characterized by the number of labels and the number of examples. The number of labels are consistent in all settings. In NLI task, each multi label example contains 10 labels, and in UFET task, each multi label example contains 2 labels. For completeness, we also provide original training data configurations.} 
    \label{tab:data_setting}
\end{table*}
We propose a set up where we distribute annotation label budget \textbf{unevenly} across training examples, resulting in unlabeled examples, single label examples, and multi label examples. We do not collect any new annotations in this work, and re-use dataset from prior work~\cite{choi2018ultra,Chen2020UncertainNL} by resplitting existing datasets to simulate different label distribution scenarios. For each task, we study \single setting, which consider a fixed number of supervised, single label example. Then, we introduce \single + \multi setting, which includes multi label examples and single label examples (but fixing the amount of total annotation same as the \single setting). Lastly, we study adding unlabeled examples \unlabel to both settings.

\subsection{Task}
We consider two classification tasks, Natural Language Inference (NLI) and fine-grained entity typing. Recent papers~\cite{Pavlick2019InherentDI,nie2020can} have shown that human annotators disagree on NLI task for its inherent ambiguity. Such disagreement is not an annotation artifact but rather exhibits the judgement of annotators with differing interpretations of entailment~\cite{Reidsma2008ExploitingA}.

Named entity recognition~\cite{Sang2003IntroductionTT}, in its vanilla setting with a handful of classes, is a straightforwad task with high inter-annotator agreement. 
However, when the label set grows, comprehensive annotation becomes challenging and most distant supervision examples only offers partial labels. Many real world tasks~\cite{Bhatia16} involve such complex large label space, where comprehensively annotating examples are often infeasible. We choose ultra-fine entity typing dataset~\cite{choi2018ultra} which provides typing into a rich ontology consisting of over 10K label candidates. Unlike NLI task, fine grained entity typing is a multi class classification task, where a single example is assigned to a \textbf{set of gold type labels}. Thus, acquiring multiple labels for the same example provides correlation among the labels (e.g., musicians are also artists).  

Table~\ref{tab:each_example} shows an example  of each task, and Table~\ref{tab:data_setting} shows full experimental data configuration, which will be explained below. 
\paragraph{NLI: Label Distribution Estimation} 
NLI is a task ~\cite{Dagan2005ThePR,Bowman2015ALA} that involves deciding whether a hypothesis \textit{h} is supported by a given premise \textit{p}. It is a three-way classification task with ``entailment", ``contradiction", and ``neutral" as labels, and recently reframed as a human label distribution prediction task.

We use the training data from the original SNLI~\cite{Bowman2015ALA} and MNLI dataset~\cite{Williams2018ABC}, containing 549K and 392K instances respectively. Recent work presents ChaosNLI dataset~\cite{nie2020can}, which collects 100 labels per example in the original SNLI/MNLI development set, (1,514 examples for SNLI, 1,599 examples for MNLI).\footnote{It covers SNLI, MNLI, and $\alpha$NLI~\cite{bhagavatula2020abductive}, and we focus our study on the first two datasets as they show more disagreement among the annotators.}

For multi label data, we use ChaosNLI dataset to sample multi-annotation examples for SNLI and MNLI. We randomly sampled 500 examples from ChasSNLI and ChasMNLI respectively for evaluation set and use the rest of ChaosNLI for training.\footnote{The original datasets split data such that premise does not occur in both train and evaluation set. This random re-partition breaks that assumption, now a premise can occur in both training and evaluation with different hypotheses. However, we find that the performance on examples with/without overlapping premise in the training set does not vary significantly.} 
For ChaosNLI in the training, We randomly sample 10 out of 100 annotations for each examples in the training set.
For single label data, we directly sample from the original SNLI/MNLI data based on the annotation budget such as 150k or 6k examples.

\paragraph{Ultra Fine Entity Typing (UFET): Multi Label Classification} UFET takes a sentence and an entity mention, and labels this mention with a set of entity types from the rich type ontology covering 10K types. 
Each example is annotated with average 5 labels: 0.9 general types
, 0.6 fine-grained types 
, and 3.9 ultra-fine types. We consider each positive \textbf{type} annotation as a single label, thus original data setting is a combination of \single and \multi examples (most of them are \multin). We simulate \single setting and \single + \multi setting for our study. 

The dataset consists of 6K crowd-sourced examples, randomly split evenly into train, development, and test sets. We fix the total number of training label budget as 500 labels. For \single setting, we randomly sample 500 examples and sample one label for each example. For \single + \multi setting, we sample 100 examples with one label, and 200 examples with two labels. We only modify training data and use the original evaluation dataset. 

\section{Learning}
We introduce learning algorithms that can handle different types of training data. We describe feature extractors for both tasks, which maps natural language to a dense vector representation $x$ then discuss learning algorithms. In the learning algorithms, we first discuss learning with annotated examples only (single label and multi label) and describe learning strategy to integrate unlabeled data. All learning configurations are optimized with the cross entropy (CE) loss. 
\subsection{Base Model}\label{sec:base_model}
We present base models at here which is used to derive input feature vector $x$ from natural language examples. Training details and hyperparameter settings can be found in the appendix. 

\paragraph{NLI} We use RoBERTa~\cite{Liu2019RoBERTaAR} based classification model, i.e., encoding concatenated hypothesis and premise and pass the resulting $[$CLS$]$ representation through a fully connected layer to predict the label distribution.
\paragraph{UFET} We follow the baseline architecture presented in \citet{choi2018ultra}, a bidirectional LSTM which generates contextualized representation. The model computes weighted sum of contextualized representation for each word in the sentence to represent an example using attention. Then this representation is used to decide the membership of each label in 10K ontology.

\subsection{Labeled Examples Only}
Several learning settings are introduced here where model only learns from labeled examples (single and multi label) disregarding unlabeled data.

\paragraph{Combined Training Set: CE (combined)}
We shuffle single and multi labeled example sets together, and train the model with this combined set.

\paragraph{Upsampling: CE (upsampling)} When we have fewer multi label examples, we upsample multi label data, to match single label data. 

\paragraph{Curriculum Learning: CE (\single then \multin) } We first train with single label data, where we often have abundant examples. Then we further fine-tune this model with multi-annotated data.

\paragraph{MixUp } 
Recent work proposed MixUp ~\cite{zhang2018mixup}, a data augmentation method that encourages the model to behave linearly in-between labeled training examples for image data. \citet{berthelot2019mixmatch} extended to interpolate between the label and unlabeled data (after assigning a psuedo labels for them). \citet{Chen2020MixTextLI} applied the MixUp to text classification tasks, showing MixUp outperforms other data augmentation techniques such as back translation~\cite{Sennrich2016ImprovingNM, zhang-etal-2021-knowing} and word replacement. We describe original MixUp algorithm below.

Given two examples $(x_m, y_m)$ and $(x_n, y_n)$, where $x$ is raw input vector and $y$ is one-hot label encoding, it constructs augmented training examples by incorporating the intuition that linear
interpolations of feature vectors should lead to linear interpolations of the associated targets:

\begin{equation*}
    \begin{array}{l}
\tilde{{x}}=\text{mix}({x_m}, x_n)= \lambda {x}_{m}+(1-\lambda) {x}_{n} \\
\tilde{{y}}=\text{mix}({y_m}, y_n)=\lambda {y}_{m}+(1-\lambda) {y}_{n}, \\
\end{array}
\end{equation*}

where $\lambda$ is a scalar hyperparameter for mixing both the inputs and labels. It is sampled from a Beta$(\eta, \eta)$ distribution with a hyper-parameter $\eta$. The newly generated training data $(\tilde{{x}}, \tilde{{y}})$ are used as a training example, and the learning objective is:

\begin{equation*}
L_{\text{mixup}} = \mathcal{L}(\tilde{{y}}, d(\tilde{{x}}, \theta))
,\end{equation*}
where $\mathcal{L}$ is the cross entropy loss and  $d(.;\phi)$ is a classifier on top of the encoder model which take the mixed representation $\tilde{{x}}$ as input and returns a probability over a label set. Interpolated annotated data $x_m$ and $x_n$ can be either single label data or multi label data. We define the loss from interpolating single label example and multi label example as $L_{s,m}$, the loss from interpolating multi label example and multi label example as $L_{m, m}$, the loss from interpolating single label example and single label example as $L_{s, s}$. Thus the MixUp \cite{zhang2018mixup} loss, in our \single + \multi setting, is defined as
\begin{equation*}
\text{Mixup}(\mathbf{X}_{s}, \mathbf{X}_{m})= L_{s, s} + L_{m, m}  + \alpha (L_{s, m}),
\end{equation*}
where $\alpha$ is a coefficient  \cite{tarvainen2017mean, berthelot2019mixmatch, fan2020bayesian}.

\subsection{Semi-supervised Learning}
Now we introduce unlabeled examples into training algorithm. Following prior work~\cite{berthelot2019mixmatch}, we generate pseudo labels for each unlabeled example. For unlabeled $x_u$, we use hidden states of the model's prediction to generate the pesudo labels~\cite{Xie2020UnsupervisedDA}. Considering the unlabeled data set $\mathbf{X}_{u} = (x_u^{1} \ldots, x_u^{n})$ where $n \in \{1 \ldots N\}$, the classifier model generates a pseudo label distribution $q^{n}$ for each data point $x_u^{n}$. We sharpen this distribution by taking the argmax of distribution $q^{n}$, making a one hot vector $\hat{q^{n}}$ over the labels. The classifier used to generate the pseudo labels trained jointly in a single end-to-end learning, using the learning signals from the labeled data.  

\paragraph{MixUp Three Types of Data}
After generating the pseduo labels for unlabeled data, we have three types of input: single label examples $\mathbf{X}_{s}$, multi label examples $\mathbf{X}_{m}$, and unlabeled examples $\mathbf{X}_{u}$, all with corresponding labels. We introduce MixUp interpolation among three types of data, integrating all into the objective function as below:
\begin{equation*}
\begin{aligned}
\text{Mixup}(\mathbf{X}_{s}, \mathbf{X}_{m}, \mathbf{X}_{u})&= L_{s, s} + L_{m, m}  \\
& + \alpha (L_{s, m} +  L_{s, u} + L_{m, u}).
\end{aligned}
\end{equation*}
For all settings, we set the maximum value of loss weight $\alpha$ as $2.0$ and linearly ramp up $\alpha$ from $0$ to its maximum value over the first $100$ iterations of training as is common practice \cite{tarvainen2017mean, berthelot2019mixmatch}.

\section{Experiments}

We present performances of our labeling scheme and learning framework in this section. All experimental results are rerun three times with different random seeds to determine the variance, which is small.\footnote{The standard deviation value of KL on all method / dataset pairs is lower than 0.02 and the standard deviation of F1 is lower than 0.01. }

\begin{table*}
\centering
\footnotesize
    \centering
    \begin{tabular}{l|l|r|r|r|r|r}
\toprule
 &   \multicolumn{3}{c}{ChaosSNLI } & \multicolumn{3}{|c}{ChaosMNLI}  \\ 
   & JSD$\downarrow$  & KL $\downarrow$ &  acc (old/new)$\uparrow$  &JSD $\downarrow$ & KL $\downarrow$ &  acc (old/new) $\uparrow$   \\ \midrule

\textcolor{gray}{\single (all)}   & \textcolor{gray}{0.229} & \textcolor{gray}{0.505}& \textcolor{gray}{0.727} / \textcolor{gray}{0.754} &   \textcolor{gray}{0.307} & \textcolor{gray}{0.781}& \textcolor{gray}{0.639} / \textcolor{gray}{0.592}  \\ \midrule 

\single (our reimpl.,subset) & 0.242&0.548&	0.684 / 0.710&	 0.308 & 0.799 & \textbf{0.670} / 0.604  \\
\single + \multi (\single then \multin) &\textbf{0.183}&\textbf{	0.211}&	\textbf{0.698 / 0.748}& \textbf{0.192} & \textbf{0.180} & 0.646 / \textbf{0.691} \\ \bottomrule
    \end{tabular}
    \caption{Results on ChaosNLI datasets in a high label budget setting. The top block results are from \citet{nie2020can}, and the row in grey color are not strictly comparable due to different evaluation sets. Single (our reimpl.,subset) is our implementation of \citet{nie2020can} and evaluate the results on the 500 examples evaluation set sampled from ChaosSNLI and ChaosMNLI.  }
    \label{tab:high_resource}
\end{table*}

\begin{table*}
\centering
\footnotesize
    \centering
   {\begin{tabular}{l|l|r|r|r|r|r|r}
\toprule
\multirow{3}{*}{Data} & \multirow{3}{*}{Learning} & \multicolumn{6}{c}{Number of Total Labels}\\
\cline{3-8}
 &   & \multicolumn{3}{c|}{150k} &  \multicolumn{3}{c}{6k} \\ 
  &   &  JSD $\downarrow$ & KL $\downarrow$ &  acc (old/new) $\uparrow$   & JSD$\downarrow$  & KL $\downarrow$ &  acc (old/new)$\uparrow$ \\ \midrule 
\single & CE&  0.312	&0.572&	0.628 / 0.578  & 0.330 & 0.753 & 0.516 / 0.526\\ 
\single & \mxone{s} & 0.300 & 0.567 & 0.628 / 0.580 & 0.321 & 0.696 & 0.518 / 0.528\\ 
\single +  \multi &CE (combined)  & 0.256 & 0.370 & 0.626 / 0.584  & 0.302 & 0.422 & 0.520 / 0.532\\
 \single + \multi & CE (upsampling) & 0.249 & 0.293 & 0.614 / 0.610  & 0.285 & 0.421 & 0.506 / 0.528 \\
 \single + \multi &CE (\single then \multin) &  \textbf{0.213} & \textbf{0.216} & \textbf{0.638 / 0.646}  & 0.298 & 0.414 & 0.519 / 0.531\\

 \single + \multi & \mxtwo{s}{m} &0.243& 0.288 &	0.598 / 0.602 & 0.271 & 0.409 & 0.520 / 0.539\\ \midrule
\single + \unlabel & \mxtwo{s}{u} & 0.294 &0.537&	0.626 / 0.566 & 0.309 & 0.617 & 0.519 / 0.529 \\

\single + \multi + \unlabel &\mxtwo{s}{u} then \multi & 0.290 & 0.510 & 0.626 / 0.570  & 0.295 & 0.571 & 0.521 / 0.533 \\
\single + \multi + \unlabel&\mxthree{s}{m}{u}  &  0.241& 0.287 & 0.596 / 0.610  & \textbf{0.266} & \textbf{0.384} & \textbf{0.522 / 0.540}\\
 \bottomrule
    \end{tabular}}
    \caption{Results on the ChaosMNLI datasets under limited annotation budget (150K, 6K). Each column block shows the number of total training annotations. All results use the same amount of annotations, and each row block uses roughly same amount of training examples (bottom row block incorporates large unlabeled data). CE represents cross entropy.  }
    \label{tab:mnli_result}
\end{table*}
\subsection{Evaluation Metrics}

\paragraph{NLI} We follow evaluation metrics from original papers~\cite{Bowman2015ALA,nie2020can}. We report classification accuracy, which is computed twice, once against aggregated gold labels in the original 5-way annotated dataset (old), and against the aggregated label from 100-way annotated dataset (new). Distributional evaluation metrics, Jensen-Shannon Divergence~\cite{endres2003new}, and Kullback-Leibler Divergence~\cite{kullback1951information} are also reported. We present analysis on different evaluation metrics in Section~\ref{sec:smoothing}. 
\paragraph{UFET} We compute macro-averaged precision, recall, and F1, and the average mean reciprocal rank (MRR), following prior work.

\subsection{NLI Results}

In Table~\ref{tab:high_resource}, we evaluate the impact of introducing multi label datasets in the full data setting. Even with a large annotation budget, learning with single label data shows a limited performance, and we see substantial gains on both accuracy and distribution metrics by replacing 5K single label examples with a small amount of multi label data (500 examples).
\single + \multi outperforms previously published results ($X_s$) from \citet{nie2020can}. Here we try vanilla curriculum learning, which first trains a model with \single data and then fine tune with \multi data.

With this encouraging initial results, we further explore different learning objectives in more constrained annotation budget scenarios (150K and 6K). The results on ChaosMNLI dataset is presented in Table~\ref{tab:mnli_result}.\footnote{The results on ChaosSNLI dataset can be found in appendix Table~\ref{tab:snli_fullresult}. It shows the same trends as the results on ChaosMNLI dataset.} Across all settings, having only single label data results in inferior performances compared to dedicating even a small amount of budget to generate multi annotated data (500 examples, each 10-way annotated).

Now we compare different methods to integrate multi label data and single label data. As a baseline, we notate simply combined multi label and single label data as CE (combined). Simple combination does not work when the number of multi label data (0.5K) is much smaller than the total number of single label data (145K), but shows comparable performance in 6K setting where multi label and single label data are more balanced (0.5K multi label data vs. 1K single label data). Upsampling multi label data shows improvement over the CE combined. CE (\single then \multin) which is first training the model with single label data and then fine tune with multi label data works better, consistently achieving strong performances in different experimental settings.

Next, we discuss gains from using MixUp data augmentation methods. We observe small yet consistent gains from using example MixUp in single label setting (i.e., \single: \mxone{s} vs. \single : CE)
confirming findings from the previous studies~\cite{zhang2018mixup}. Integrating multi label training examples into MixUp objective shows gains in low annotation budget setting. In high annotation budget settings, where we have fewer multi label examples (500 multi vs. 145K single), CE (\single then \multin) yields better results. Nonetheless, MixUp augmentation shows consistent gains compared to shuffling (MixUp(\singlen, \multin) vs. CE(combined)).

Our results suggest that annotation budget should be distributed carefully. Even under same label budget and the same learning objective, distribution of labels among examples resulted in performance differences (i.e., \single: CE vs. \single + \multi: CE (combined)). Incorporating unlabeled examples (\mxtwo{s}{u} vs \mxone{s}) improves the performances in low label budget settings (6K), but is detrimental in high label budget settings (150K). We hypothesize that imperfect pseudo label for unlabeled examples can interfere the learning. 
\begin{table*}
\centering
\footnotesize
    \centering
    \resizebox{2\columnwidth}{!}{
    \begin{tabular}{l|l|l|r|r|r|r|r|r|r}
\toprule
\multirow{2}{*}{Data} & \multirow{2}{*}{Learning} &   \multicolumn{4}{c}{Development Set} & \multicolumn{4}{|c}{Test Set}  \\ 
\cline{3-6} \cline{7-10}
& & MRR  & P &  R & F1 & MRR  &  P & R & F1  \\ 
\midrule
\citet{choi2018ultra} (w / full crowd data) & CE & 0.181 &46.2& 15.7& 23.4 & 0.178 & 44.7 & 15.3 & 22.8 \\
  \citet{choi2018ultra} (w / full crowd data)& \mxtwo{s}{m}&\textbf{0.197} &\textbf{46.4} &	\textbf{19.7}& \textbf{27.7} & \textbf{0.198} & \textbf{45.3} & \textbf{20.3} & \textbf{28.0} \\
 \hline \hline \\ [-2.0ex]
 
 \single& CE & 0.172 &45.1&	9.1 & 15.2 & 0.172 & 45.8 & 9.3 & 15.4\\
 \single& \mxone{s} & 0.174 & 45.5 & 9.2 & 15.4 & 0.176 & 46.0 & 9.5 & 15.7 \\
 \single + \multi & CE (combined) & 0.177 & 45.6 & 10.0 & 16.4 & 0.180 & 46.1 & 10.3 & 16.8\\
 \single + \multi  & CE (\single then \multin) & 0.179 & 46.2 & 9.9 & 16.3 & 0.181 & 48.5 & 10.1 & 16.7 \\
  \single + \multi  & \mxtwo{s}{m}& \textbf{0.181} & \textbf{48.7} & 10.2 & 16.9 & \textbf{0.183} & \textbf{49.6} & 10.3 & 17.1\\
 \midrule
\single + \unlabel & \mxtwo{s}{u} & 0.172 &47.0&	9.5 & 15.8 & 0.173 & 47.4 & 9.6 & 16.0\\ 
\single + \multi + \unlabel & \mxthree{s}{m}{u} &0.180 & 48.5 & \textbf{10.6} & \textbf{17.4} & 0.181 & 49.1 & \textbf{10.6} & \textbf{17.4}\\
 \bottomrule
    \end{tabular}}
    \caption{Results on UFET dataset. Top two rows use the full crowd-sourced data and the bottom rows are based on smaller label annotation budgets, thus results are not comparable (see Table~\ref{tab:data_setting} for details).}
    \label{tab:entity_main_result}
\end{table*}
\subsection{UFET Results} \label{sec:entity_result}
Table \ref{tab:entity_main_result} reports performances on ultra fine entity typing dataset. Instead of using both crowd-sourced data and distant supervision data \cite{choi2018ultra}, we focus on crowd-sourced data to simulate single label and multi label settings. Similar to previous results, each row block represents different annotation label budgets. Top two rows use the full crowd-sourced data and the results are not comparable to the bottom rows. The bottom rows are based on different annotation budgets such as 500 single label data (see Table~\ref{tab:data_setting} for details). Again in this task, using a single label per example results in inferior performances compared to having multiple labels per example (\single + \multi: CE (\single then \multin) vs. \single : CE ) as multi label data helps model to learn label-label interaction. Similar to NLI task, adding MixUp objective to the single label setting shows gains (\single: \mxone{s} vs. \single : CE). Having multi label data is crucial for high performances, and MixUp again shows gains in this low resource setting. 


\subsection{Analysis }
\textbf{How does different learning algorithm compares under domain shift?}
We compare two promising methods -- single and then multi (CE (\single then \multin)) and MixUp (\mxtwo{s}{m}) for their performance in out of domain setting. Prior work suggested MixUp approaches can effectively compensate for the mixmatch between test data and training data~\cite{zhu2019mixup}. 
Table~\ref{tab:ood} shows the performances of model trained on SNLI and tested on MNLI dataset. We observe improved accuracy with MixUp compared to training with the curriculum approach (train with single label data and then fine tuning with multi label data).
\begin{table} 
\footnotesize
    \centering
    \resizebox{1\columnwidth}{!}{\begin{tabular}{l|l|l|l}
\toprule
    Learning & JSD & KL &  acc (old/new)\\ \midrule
 CE (\single then \multin)& 0.339 & {\bf 0.479} & 0.432 / 0.324\\
\mxtwo{s}{m}& {\bf 0.324} & 0.489 & {\bf 0.490 / 0.480} \\
 \bottomrule
    \end{tabular}}
    \caption{Out of domain evaluation results: trained on SNLI dataset and tested on MNLI dataset. All used the same amount of annotation (6K labeled data).}
    \label{tab:ood}
\end{table}

\paragraph{Should we carefully select which examples to have multiple annotations?} Maybe. 
We experiment on how to select examples to have multiple annotations, using the ideas from ~\citet{Swayamdipta2020DatasetCM}. We finetune with 1K most hard-to-learn, most easy-to-learn, most ambiguous, and randomly sampled examples. Easy-to-learn examples, with lowest label distribution entropy, are the least effective, but the difference is small in our settings. Similarly, our experiments of changing the number of labels (5-way, 10-way, 20-way) did not result in meaningful differences. The experimental results can be found in Table \ref{tab:label_count_comparison} in the appendix.

\paragraph{Can we use multi label data exclusively without any single label data?} In our main experiments, we mixed multi label data with single label data. Here we present a study comparing a setting with $X_m$ only and $X_s$ only on the NLI task, while keeping \textbf{small} annotation budget steady (1K labels). On ChaosSNLI dataset, the model trained with single label data (1000 examples, 1-way annotated) achieves JSD: 0.3578, KL: 0.4671, and acc (old/new): 0.581/0.602. For multi label data (500 examples, 2-way annotated), we get JSD: 0.3355, KL: 0.4529, and acc (old/new):  0.592/0.614. We observe a similar trend for ChaosMNLI dataset as well. We cannot claim that $X_m$ only will outperform $X_s$ only in all settings -- as models will benefit from being exposed to diverse examples, but in this low resource setting, we observe gains from using multi annotated data alone.
\subsection{Calibration: Alternative Approach to Improve Label Distribution Prediction }\label{sec:smoothing}
We introduce using multi label training examples as an efficient way to estimate the distribution of labels. Here, we provide a study of alternative ways to improve label distribution prediction, borrowing ideas from calibration literature, and compare the calibration with training with multi label data. 

The key observation is that the predicted label distribution from model trained with single label was over confident, with smaller predicted label entropy $0.414$ in Table \ref{tab:smoothing} compared to the human annotated label entropy $0.732$. Thus, we smooth the output distribution with three calibration methods~\cite{Guo2017OnCO,Miller1996AGO}. The temp. scaling and pred smoothing are post-hoc and do not require re-training of the model. For all methods, we tuned a single scalar hyperparameter per dataset such that the entropy of prediction label distribution matching the entropy of human label distribution. 

\begin{table}
\centering
\footnotesize
    \resizebox{.49\textwidth}{!}{
    \begin{tabular}{l|l|l|l|l}
\toprule
   &  JSD & KL  &  acc (old/new)  & $H$  \\ \midrule

\single  & 	  0.308 & 0.799 & 0.670 / 0.604 & 0.414 \\
+ temp. scaling &  0.233 & 0.324 & 0.670 / 0.604 & 0.720 \\
+ pred smoothing&  0.245 & 0.347 & 0.670 / 0.604 & 0.722\\ 
+ train smoothing &  0.252 & 0.372 & \textbf{0.680} / 0.602 & 0.701\\
\single then \multi  & \textbf{0.192} & \textbf{0.180} & 0.646 / \textbf{0.691} & 0.868 \\ \bottomrule
    \end{tabular}}
    \caption{Results on ChaosMNLI dataset  with calibration methods. The entropy value of human label distribution for ChaosMNLI is $0.732$. $H$ represents the predicted label entropy. 
    Lower entropy indicates higher confidence.
    }
    \label{tab:smoothing}
\end{table}

\setlist{nolistsep}
\begin{itemize}[noitemsep]
\item \textbf{temp. scaling}: scaling by multiplying non-normalized logits by a scalar hyperparameter.
\item \textbf{pred smoothing}: process softmaxed label distribution by moving $\alpha$ probability mass from the label with the highest mass to the all labels equally.
\item \textbf{train smoothing}: process training label distribution by shifting $\alpha$ probability mass from the gold label to the all labels equally.
\end{itemize}
Table~\ref{tab:smoothing} reports performances of calibration methods. We find all calibration methods improve performance on both distribution metrics (JSD and KL). Temperature scaling yields slightly better results than label smoothing, consistent with the findings from ~\citet{Desai2020CalibrationOP} which shows temperature scaling is better for in-domain calibration compared to label smoothing. Nonetheless, all these results were substantially worse than using multi label data during the training.

\begin{figure}
\centering
\footnotesize
\setlength{\tabcolsep}{1.5pt}
\begin{tabular}{cc}
~~~(a) Human label  & (b)  \single  model  \\
\includegraphics[width=0.22\textwidth]{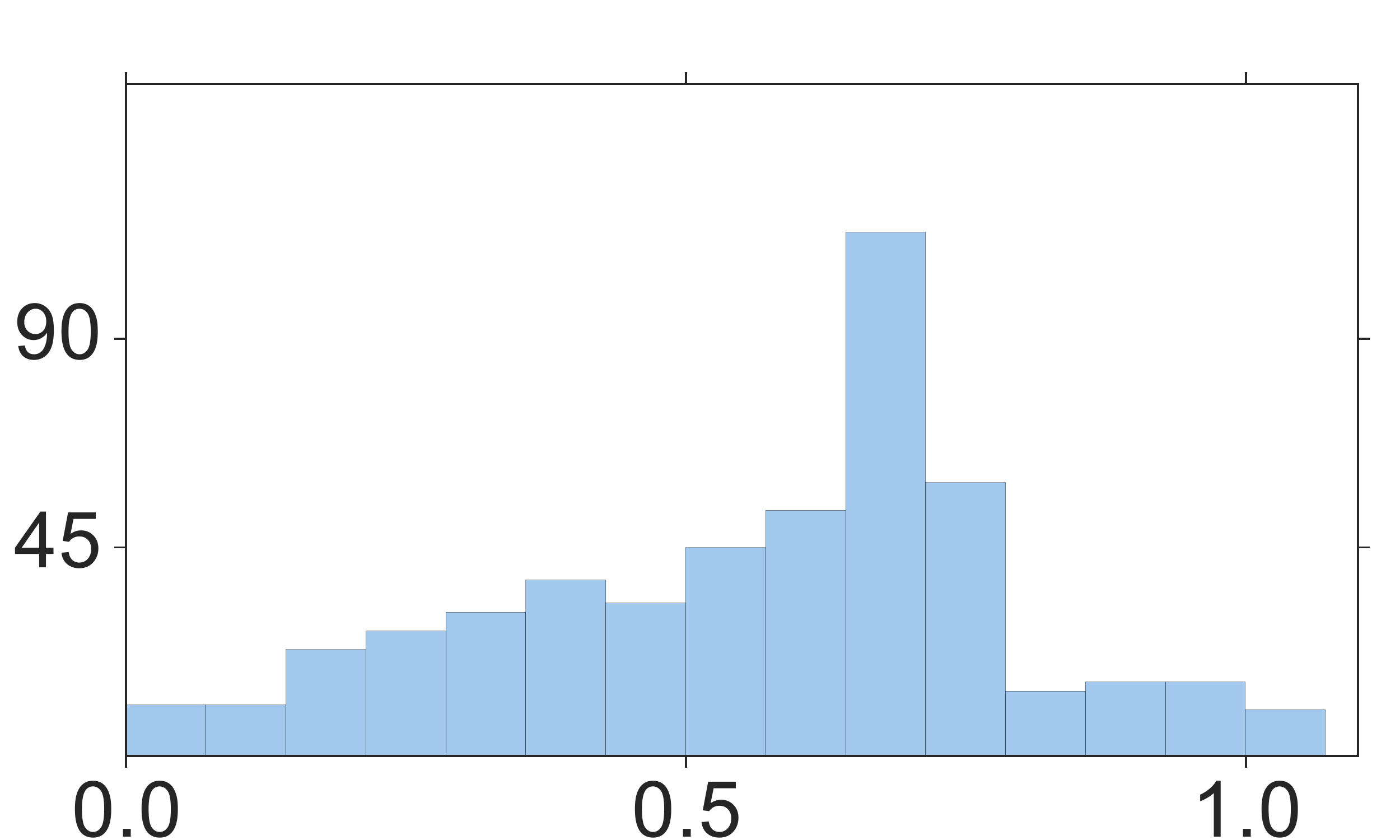}
& 
\includegraphics[width=0.23\textwidth]{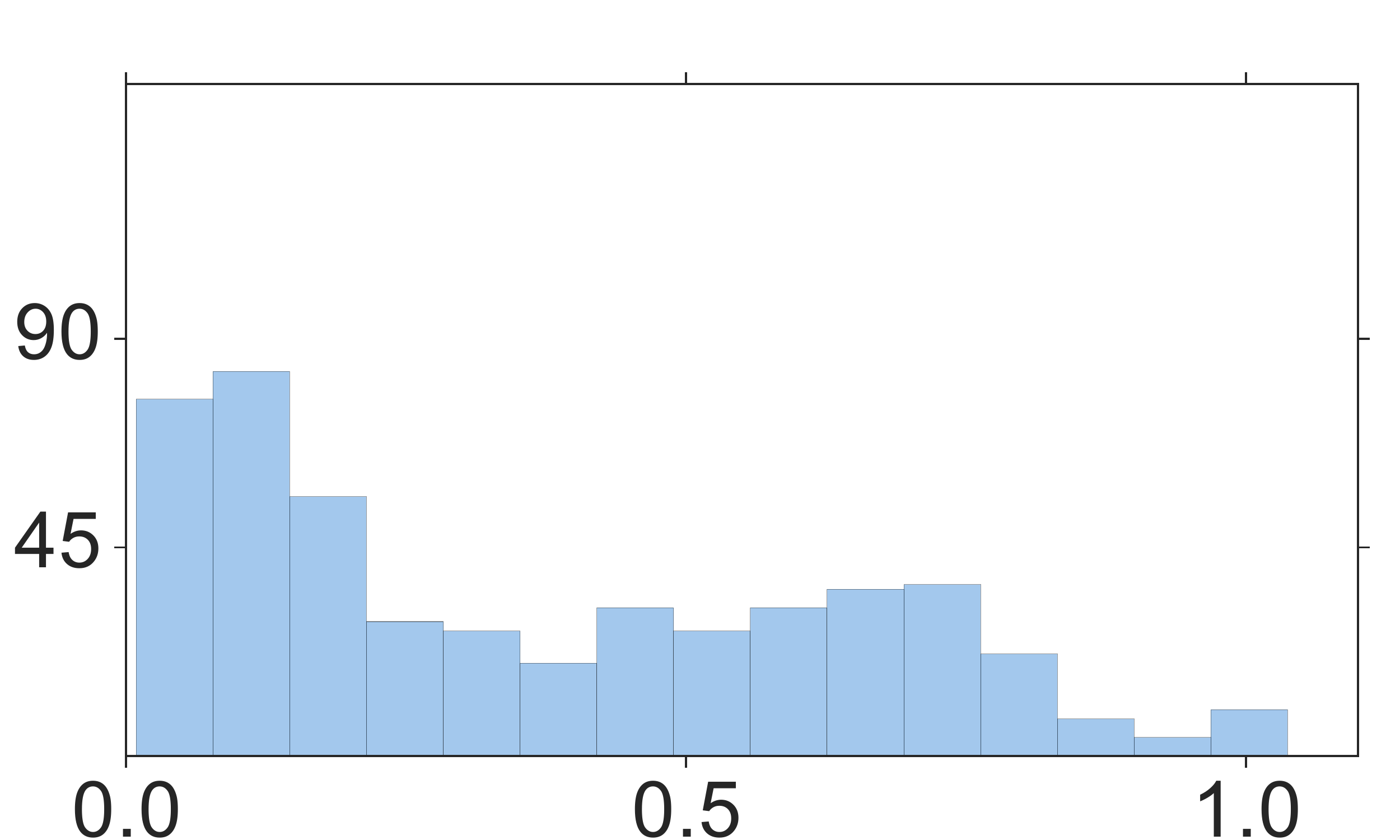}~~~~ \\
(c) Calibrated \single model  & (d) \single then \multi model \\
\includegraphics[width=0.23\textwidth]{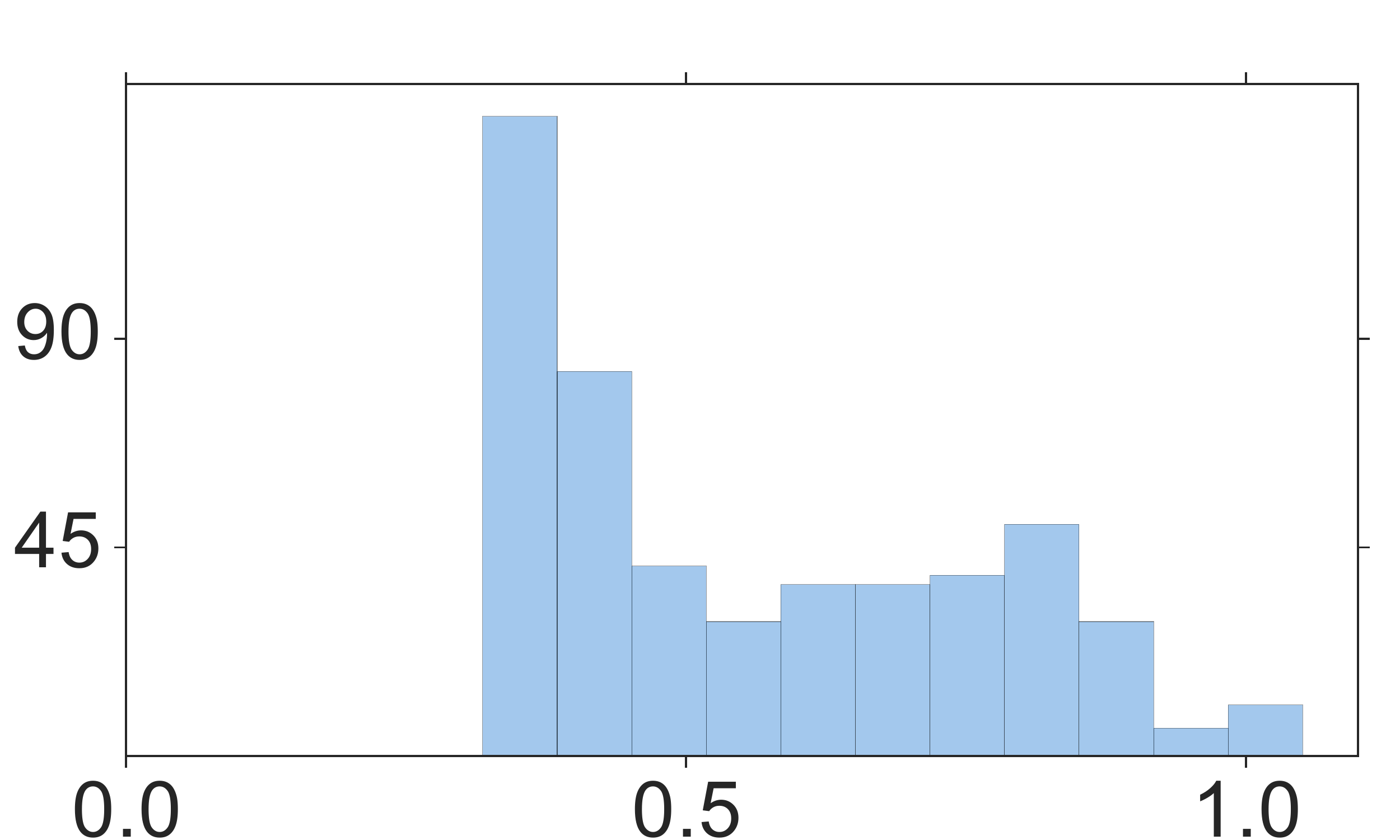} &
\includegraphics[width=0.23\textwidth]{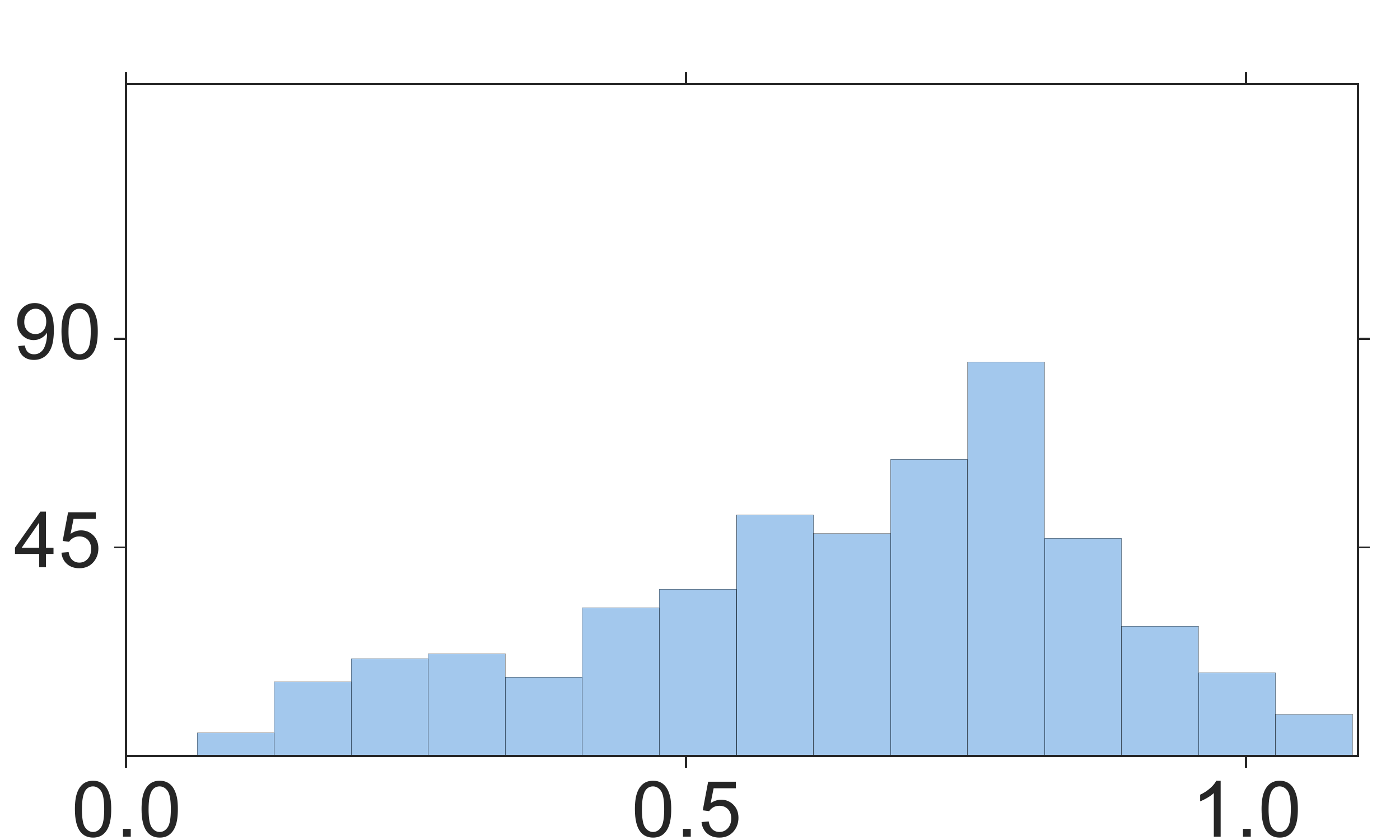}~~~~
\\
\end{tabular}
\caption{The empirical distribution of label/prediction entropy on ChaosSNLI dataset, where
x-axis denotes the entropy value and y-axis denotes the example count on the entropy bin.
Initial model prediction (b) shows low entropy values for many examples, being over-confident. Post-hoc calibration nicely shifts the distribution to be less confident, but with artifacts of not being confident on any examples. Finetuning on the small amount of multi-annotated data in (d) successfully simulate the entropy distribution of human labels in (a).}
\label{fig:labelentropy}
\end{figure}
\paragraph{Can we estimate the distribution of ambiguous and less ambiguous examples?} Figure~\ref{fig:labelentropy} shows the empirical example distribution over entropy bins. The leftmost plot (a) shows the annotated human label entropy over our evaluation set, and the plot (b) next to it shows the prediction entropy of the baseline RoBERTa model predictions. The model is over-confident about its prediction with single label examples. 
With label smoothing (plot c), the over-confidence problem is relieved, but the entropy distribution still does not match the distribution of ground truth. Training with multi label data (plot d) makes the prediction distribution similar to the ground truth. 

\section{Related Work}
Assessing the annotation cost associated with learning has long been studied~\cite{Turney2002TypesOC}. \citet{Sheng2008GetAL} studies the tradeoff between collecting multiple labels per example vs. annotating more examples. Researchers have also explored different data labeling strategies, such as active learning~\cite{fang-etal-2017-learning}, providing fine-grained rationales~\cite{Dua2020BenefitsOI}, retrospectively studying the amount of training data necessary for generalization~\cite{mishra2020we}, and the policy learning approach ~\cite{kratzwald2020learning}. In this work, we study uneven distribution of label annotation budget for training examples, which has not been explored to our knowledge.

Label propagation has been extensively used to infer pseudo-labels for unlabeled data, which are used to train the classifier~\cite{zhou2004learning,li2016two}. 
Our use of MixUp can be viewed as a way to propagate label information between the single labeled, multi labeled, and unlabeled data.

Rich prior work studies ambiguity in language interpretations~\cite{Aroyo2015TruthIA}. A few studies~\cite{Passonneau2012MultiplicityAW,ferracane-etal-2021-answer} frame diverging, subjective interpretations as a multi label classification, and few studies~\cite{Glickman2005APC,Zhang2017OrdinalCI,Chen2020UncertainNL} introduce graded human responses. \citet{Mayhew2020SimultaneousTA} studies training machine translation system with the goal of generating diverse set of reference translations.  ~\citet{Pavlick2019InherentDI} examines the distribution behind human references for NLI and ~\citet{nie2020can} presents a larger-scale data collection that we build on.

Earlier version of this paper \cite{zhang2021capturing} study capturing inherent human disagreement
in the NLI task through calibration and using a
small amount of multi-annotated training examples. 
This paper expands upon it, introducing a new learning framework for such uneven label distribution schemes. Concurrent to our work, ~\citet{zhou2021distributed} introduces distributed NLI, a new NLU task with a goal to predict the distribution of human judgements by applying additional distribution estimation methods such as Monte Carlo (MC) Dropout and deep ensemble methods. While we share a similar goal, our work focuses on how to distribute training labels across examples and how to learn under this new label distribution scheme. 

\section{Conclusion}
Our work demonstrates the benefits from introducing a small amount of multi label examples at the cost of annotating fewer examples. The proposed learning algorithm, extended from MixUp, flexibly takes signals from different types of training examples (single label data, multi label data, and unlabeled data) and show gains upon simply combining different datasets in low annotation budget settings. In this work, we retrospectively study with existing data to question original annotation collection designs. 
Exploring reinforcement learning or active learning to predict an optimal distribution of annotation budget will be an exciting avenue for future work.
\newpage

\newpage

\section{Acknowledgements}
The authors thank Greg Durrett, Raymond Mooney, Kaj Bostrom, Yasumasa Onoe, and Kenton Lee for helpful comments on the paper draft. 

\bibliography{custom}

\begin{thebibliography}{53}
\expandafter\ifx\csname natexlab\endcsname\relax\def\natexlab#1{#1}\fi

\bibitem[{Aroyo and Welty(2015)}]{Aroyo2015TruthIA}
Lora Aroyo and Chris Welty. 2015.
\newblock Truth is a lie: Crowd truth and the seven myths of human annotation.
\newblock \emph{AI Mag.}, 36:15--24.

\bibitem[{Asher and Lascarides(2005)}]{Asher2005LogicsOC}
Nicholas Asher and A.~Lascarides. 2005.
\newblock Logics of conversation.
\newblock In \emph{Studies in natural language processing}.

\bibitem[{Berthelot et~al.(2019)Berthelot, Carlini, Goodfellow, Papernot,
  Oliver, and Raffel}]{berthelot2019mixmatch}
David Berthelot, Nicholas Carlini, Ian Goodfellow, Nicolas Papernot, Avital
  Oliver, and Colin Raffel. 2019.
\newblock Mixmatch: A holistic approach to semi-supervised learning.
\newblock \emph{arXiv preprint arXiv:1905.02249}.

\bibitem[{Bhagavatula et~al.(2020)Bhagavatula, Le~Bras, Malaviya, Sakaguchi,
  Holtzman, Rashkin, Downey, Yih, and Choi}]{bhagavatula2020abductive}
Chandra Bhagavatula, Ronan Le~Bras, Chaitanya Malaviya, Keisuke Sakaguchi, Ari
  Holtzman, Hannah Rashkin, Doug Downey, Wen-tau Yih, and Yejin Choi. 2020.
\newblock Abductive commonsense reasoning.
\newblock In \emph{International Conference on Learning Representations
  (ICLR)}.

\bibitem[{Bhatia et~al.(2016)Bhatia, Dahiya, Jain, Kar, Mittal, Prabhu, and
  Varma}]{Bhatia16}
K.~Bhatia, K.~Dahiya, H.~Jain, P.~Kar, A.~Mittal, Y.~Prabhu, and M.~Varma.
  2016.
\newblock \href {http://manikvarma.org/downloads/XC/XMLRepository.html} {The
  extreme classification repository: Multi-label datasets and code}.

\bibitem[{Bowman et~al.(2015)Bowman, Angeli, Potts, and
  Manning}]{Bowman2015ALA}
Samuel~R. Bowman, Gabor Angeli, Christopher Potts, and Christopher~D. Manning.
  2015.
\newblock A large annotated corpus for learning natural language inference.
\newblock \emph{Proceedings of the Conference on Empirical Methods in Natural
  Language Processing (EMNLP)}, abs/1508.05326.

\bibitem[{Chen et~al.(2020{\natexlab{a}})Chen, Yang, and
  Yang}]{Chen2020MixTextLI}
Jiaao Chen, Zichao Yang, and Diyi Yang. 2020{\natexlab{a}}.
\newblock Mixtext: Linguistically-informed interpolation of hidden space for
  semi-supervised text classification.
\newblock \emph{Proceedings of the Annual Meeting of the Association for
  Computational Linguistics (ACL)}, abs/2004.12239.

\bibitem[{Chen et~al.(2020{\natexlab{b}})Chen, Jiang, Sakaguchi, and
  Durme}]{Chen2020UncertainNL}
Tongfei Chen, Zhengping Jiang, Keisuke Sakaguchi, and Benjamin~Van Durme.
  2020{\natexlab{b}}.
\newblock Uncertain natural language inference.
\newblock In \emph{Proceedings of the Annual Meeting of the Association for
  Computational Linguistics (ACL)}.

\bibitem[{Choi et~al.(2018)Choi, Levy, Choi, and Zettlemoyer}]{choi2018ultra}
Eunsol Choi, Omer Levy, Yejin Choi, and Luke Zettlemoyer. 2018.
\newblock Ultra-fine entity typing.
\newblock In \emph{Proceedings of the Annual Meeting of the Association for
  Computational Linguistics (ACL)}, pages 87--96.

\bibitem[{Dagan et~al.(2005)Dagan, Glickman, and Magnini}]{Dagan2005ThePR}
I.~Dagan, Oren Glickman, and B.~Magnini. 2005.
\newblock The pascal recognising textual entailment challenge.
\newblock In \emph{MLCW}.

\bibitem[{Desai and Durrett(2020)}]{Desai2020CalibrationOP}
Shrey Desai and Greg Durrett. 2020.
\newblock Calibration of pre-trained transformers.
\newblock \emph{Conference on Empirical Methods in Natural Language
  Processing}, abs/2003.07892.

\bibitem[{Dodge et~al.(2019)Dodge, Gururangan, Card, Schwartz, and
  Smith}]{dodge2019show}
Jesse Dodge, Suchin Gururangan, Dallas Card, Roy Schwartz, and Noah~A Smith.
  2019.
\newblock Show your work: Improved reporting of experimental results.
\newblock In \emph{Proceedings of the 2019 Conference on Empirical Methods in
  Natural Language Processing and the 9th International Joint Conference on
  Natural Language Processing (EMNLP-IJCNLP)}, pages 2185--2194.

\bibitem[{Dua et~al.(2020)Dua, Singh, and Gardner}]{Dua2020BenefitsOI}
Dheeru Dua, Sameer Singh, and Matt Gardner. 2020.
\newblock Benefits of intermediate annotations in reading comprehension.
\newblock In \emph{Proceedings of the Annual Meeting of the Association for
  Computational Linguistics (ACL)}.

\bibitem[{Endres and Schindelin(2003)}]{endres2003new}
Dominik~Maria Endres and Johannes~E Schindelin. 2003.
\newblock A new metric for probability distributions.
\newblock \emph{IEEE Transactions on Information theory}, 49(7):1858--1860.

\bibitem[{Fan et~al.(2020)Fan, Zhang, Chen, and Zhou}]{fan2020bayesian}
Xinjie Fan, Shujian Zhang, Bo~Chen, and Mingyuan Zhou. 2020.
\newblock Bayesian attention modules.
\newblock \emph{arXiv preprint arXiv:2010.10604}.

\bibitem[{Fang et~al.(2017)Fang, Li, and Cohn}]{fang-etal-2017-learning}
Meng Fang, Yuan Li, and Trevor Cohn. 2017.
\newblock \href {https://doi.org/10.18653/v1/D17-1063} {Learning how to active
  learn: A deep reinforcement learning approach}.
\newblock In \emph{Proceedings of the 2017 Conference on Empirical Methods in
  Natural Language Processing}, pages 595--605, Copenhagen, Denmark.
  Association for Computational Linguistics.

\bibitem[{Ferracane et~al.(2021)Ferracane, Durrett, Li, and
  Erk}]{ferracane-etal-2021-answer}
Elisa Ferracane, Greg Durrett, Junyi~Jessy Li, and Katrin Erk. 2021.
\newblock \href {https://doi.org/10.18653/v1/2021.naacl-main.129} {Did they
  answer? subjective acts and intents in conversational discourse}.
\newblock In \emph{Proceedings of the 2021 Conference of the North American
  Chapter of the Association for Computational Linguistics: Human Language
  Technologies}, pages 1626--1644, Online. Association for Computational
  Linguistics.

\bibitem[{Freitag et~al.(2021)Freitag, Foster, Grangier, Ratnakar, Tan, and
  Macherey}]{Freitag2021ExpertsEA}
Markus Freitag, George~F. Foster, David Grangier, Viresh Ratnakar, Qijun Tan,
  and Wolfgang Macherey. 2021.
\newblock Experts, errors, and context: A large-scale study of human evaluation
  for machine translation.
\newblock \emph{ArXiv}, abs/2104.14478.

\bibitem[{Glickman et~al.(2005)Glickman, Dagan, and Koppel}]{Glickman2005APC}
Oren Glickman, I.~Dagan, and Moshe Koppel. 2005.
\newblock A probabilistic classification approach for lexical textual
  entailment.
\newblock In \emph{Proceedings of the {AAAI} Conference on Artificial
  Intelligence}.

\bibitem[{Guo et~al.(2018)Guo, Pleiss, Sun, and Weinberger}]{Guo2017OnCO}
Chuan Guo, Geoff Pleiss, Yu~Sun, and Kilian~Q. Weinberger. 2018.
\newblock On calibration of modern neural networks.
\newblock \emph{Proceedings of the International Conference on Machine Learning
  (ICML)}.

\bibitem[{Kingma and Ba(2014)}]{kingma2014adam}
Diederik~P Kingma and Jimmy Ba. 2014.
\newblock Adam: A method for stochastic optimization.
\newblock \emph{arXiv preprint arXiv:1412.6980}.

\bibitem[{Kratzwald et~al.(2020)Kratzwald, Feuerriegel, and
  Sun}]{kratzwald2020learning}
Bernhard Kratzwald, Stefan Feuerriegel, and Huan Sun. 2020.
\newblock Learning a cost-effective annotation policy for question answering.
\newblock In \emph{Proceedings of the 2020 Conference on Empirical Methods in
  Natural Language Processing (EMNLP)}, pages 3051--3062.

\bibitem[{Kullback and Leibler(1951)}]{kullback1951information}
Solomon Kullback and Richard~A Leibler. 1951.
\newblock On information and sufficiency.
\newblock \emph{The annals of mathematical statistics}, 22(1):79--86.

\bibitem[{Li et~al.(2016)Li, Xu, Zhang, and Zhou}]{li2016two}
Shoushan Li, Jian Xu, Dong Zhang, and Guodong Zhou. 2016.
\newblock Two-view label propagation to semi-supervised reader emotion
  classification.
\newblock In \emph{Proceedings of COLING 2016, the 26th International
  Conference on Computational Linguistics: Technical Papers}, pages 2647--2655.

\bibitem[{Liu et~al.(2021)Liu, Lee, Jia, and Liang}]{Liu2021CanSA}
Nelson~F. Liu, T.~Lee, Robin Jia, and Percy Liang. 2021.
\newblock Can small and synthetic benchmarks drive modeling innovation? a
  retrospective study of question answering modeling approaches.
\newblock \emph{ArXiv}, abs/2102.01065.

\bibitem[{Liu et~al.(2019)Liu, Ott, Goyal, Du, Joshi, Chen, Levy, Lewis,
  Zettlemoyer, and Stoyanov}]{Liu2019RoBERTaAR}
Y.~Liu, Myle Ott, Naman Goyal, Jingfei Du, Mandar Joshi, Danqi Chen, Omer Levy,
  M.~Lewis, Luke Zettlemoyer, and Veselin Stoyanov. 2019.
\newblock Roberta: A robustly optimized bert pretraining approach.
\newblock \emph{ArXiv}, abs/1907.11692.

\bibitem[{Mayhew et~al.(2020)Mayhew, Bicknell, Brust, McDowell, and
  Monroe}]{Mayhew2020SimultaneousTA}
Stephen Mayhew, K.~Bicknell, Chris Brust, Bill McDowell, and Will Monroe. 2020.
\newblock Simultaneous translation and paraphrase for language education.
\newblock In \emph{NGT@ACL}.

\bibitem[{Miller et~al.(1996)Miller, Rao, Rose, and Gersho}]{Miller1996AGO}
David~J. Miller, A.~Rao, K.~Rose, and A.~Gersho. 1996.
\newblock A global optimization technique for statistical classifier design.
\newblock \emph{IEEE Trans. Signal Process.}, 44:3108--3122.

\bibitem[{Min et~al.(2020)Min, Michael, Hajishirzi, and
  Zettlemoyer}]{Min2020AmbigQAAA}
Sewon Min, Julian Michael, Hannaneh Hajishirzi, and Luke Zettlemoyer. 2020.
\newblock Ambigqa: Answering ambiguous open-domain questions.
\newblock \emph{Proceedings of the Conference on Empirical Methods in Natural
  Language Processing (EMNLP)}, abs/2004.10645.

\bibitem[{Mishra and Sachdeva(2020)}]{mishra2020we}
Swaroop Mishra and Bhavdeep~Singh Sachdeva. 2020.
\newblock Do we need to create big datasets to learn a task?
\newblock In \emph{Proceedings of SustaiNLP: Workshop on Simple and Efficient
  Natural Language Processing}, pages 169--173.

\bibitem[{Nie et~al.(2020)Nie, Zhou, and Bansal}]{nie2020can}
Yixin Nie, Xiang Zhou, and Mohit Bansal. 2020.
\newblock What can we learn from collective human opinions on natural language
  inference data?
\newblock In \emph{Proceedings of the 2020 Conference on Empirical Methods in
  Natural Language Processing (EMNLP)}, pages 9131--9143.

\bibitem[{Passonneau et~al.(2012)Passonneau, Bhardwaj, Salleb-Aouissi, and
  Ide}]{Passonneau2012MultiplicityAW}
Rebecca~J. Passonneau, Vikas Bhardwaj, Ansaf Salleb-Aouissi, and Nancy Ide.
  2012.
\newblock Multiplicity and word sense: evaluating and learning from multiply
  labeled word sense annotations.
\newblock \emph{Language Resources and Evaluation}, 46:219--252.

\bibitem[{Pavlick and Kwiatkowski(2019)}]{Pavlick2019InherentDI}
Ellie Pavlick and Tom Kwiatkowski. 2019.
\newblock Inherent disagreements in human textual inferences.
\newblock \emph{Transactions of the Association for Computational Linguistics},
  7:677--694.

\bibitem[{Rajpurkar et~al.(2016)Rajpurkar, Zhang, Lopyrev, and
  Liang}]{Rajpurkar2016SQuAD10}
Pranav Rajpurkar, Jian Zhang, Konstantin Lopyrev, and Percy Liang. 2016.
\newblock Squad: 100, 000+ questions for machine comprehension of text.
\newblock In \emph{Proceedings of the Conference on Empirical Methods in
  Natural Language Processing (EMNLP)}.

\bibitem[{Reidsma and op~den Akker(2008)}]{Reidsma2008ExploitingA}
D.~Reidsma and Rieks op~den Akker. 2008.
\newblock Exploiting ‘subjective’ annotations.
\newblock In \emph{COLING 2008}.

\bibitem[{Sang and Meulder(2003)}]{Sang2003IntroductionTT}
E.~T.~K. Sang and F.~D. Meulder. 2003.
\newblock Introduction to the conll-2003 shared task: Language-independent
  named entity recognition.
\newblock \emph{ArXiv}, cs.CL/0306050.

\bibitem[{Sennrich et~al.(2016)Sennrich, Haddow, and
  Birch}]{Sennrich2016ImprovingNM}
Rico Sennrich, B.~Haddow, and Alexandra Birch. 2016.
\newblock Improving neural machine translation models with monolingual data.
\newblock \emph{ArXiv}, abs/1511.06709.

\bibitem[{Sheng et~al.(2008)Sheng, Provost, and Ipeirotis}]{Sheng2008GetAL}
V.~Sheng, F.~Provost, and Panagiotis~G. Ipeirotis. 2008.
\newblock Get another label? improving data quality and data mining using
  multiple, noisy labelers.

\bibitem[{Swayamdipta et~al.(2020)Swayamdipta, Schwartz, Lourie, Wang,
  Hajishirzi, Smith, and Choi}]{Swayamdipta2020DatasetCM}
Swabha Swayamdipta, Roy Schwartz, Nicholas Lourie, Yizhong Wang, Hannaneh
  Hajishirzi, Noah~A. Smith, and Yejin Choi. 2020.
\newblock Dataset cartography: Mapping and diagnosing datasets with training
  dynamics.
\newblock In \emph{Conference on Empirical Methods in Natural Language
  Processing}.

\bibitem[{Tarvainen and Valpola(2017)}]{tarvainen2017mean}
Antti Tarvainen and Harri Valpola. 2017.
\newblock Mean teachers are better role models: Weight-averaged consistency
  targets improve semi-supervised deep learning results.
\newblock In \emph{Proceedings of the 31st International Conference on Neural
  Information Processing Systems}, pages 1195--1204.

\bibitem[{Turney(2002)}]{Turney2002TypesOC}
Peter~D. Turney. 2002.
\newblock Types of cost in inductive concept learning.
\newblock \emph{ArXiv}, cs.LG/0212034.

\bibitem[{Vaswani et~al.(2017)Vaswani, Shazeer, Parmar, Uszkoreit, Jones,
  Gomez, Kaiser, and Polosukhin}]{transformer}
Ashish Vaswani, Noam Shazeer, Niki Parmar, Jakob Uszkoreit, Llion Jones,
  Aidan~N Gomez, {\L}ukasz Kaiser, and Illia Polosukhin. 2017.
\newblock Attention is all you need.
\newblock In \emph{NeurIPS}.

\bibitem[{Wang et~al.(2019)Wang, Pruksachatkun, Nangia, Singh, Michael, Hill,
  Levy, and Bowman}]{Wang2019SuperGLUEAS}
Alex Wang, Yada Pruksachatkun, Nikita Nangia, Amanpreet Singh, Julian Michael,
  Felix Hill, Omer Levy, and Samuel~R. Bowman. 2019.
\newblock Superglue: A stickier benchmark for general-purpose language
  understanding systems.
\newblock In \emph{Proceedings of Advances in Neural Information Processing
  Systems (NeurIPS)}.

\bibitem[{Williams et~al.(2018)Williams, Nangia, and Bowman}]{Williams2018ABC}
Adina Williams, Nikita Nangia, and Samuel~R. Bowman. 2018.
\newblock A broad-coverage challenge corpus for sentence understanding through
  inference.
\newblock \emph{naacl}, abs/1704.05426.

\bibitem[{Wolf et~al.(2020)Wolf, Chaumond, Debut, Sanh, Delangue, Moi, Cistac,
  Funtowicz, Davison, Shleifer et~al.}]{wolf2020transformers}
Thomas Wolf, Julien Chaumond, Lysandre Debut, Victor Sanh, Clement Delangue,
  Anthony Moi, Pierric Cistac, Morgan Funtowicz, Joe Davison, Sam Shleifer,
  et~al. 2020.
\newblock Transformers: State-of-the-art natural language processing.
\newblock In \emph{Proceedings of the 2020 Conference on Empirical Methods in
  Natural Language Processing: System Demonstrations}, pages 38--45.

\bibitem[{Xie et~al.(2020)Xie, Dai, Hovy, Luong, and
  Le}]{Xie2020UnsupervisedDA}
Qizhe Xie, Zihang Dai, E.~Hovy, Minh-Thang Luong, and Quoc~V. Le. 2020.
\newblock Unsupervised data augmentation for consistency training.
\newblock \emph{arXiv: Learning}.

\bibitem[{Zhang et~al.(2018)Zhang, Cisse, Dauphin, and
  Lopez-Paz}]{zhang2018mixup}
Hongyi Zhang, Moustapha Cisse, Yann~N. Dauphin, and David Lopez-Paz. 2018.
\newblock \href {https://openreview.net/forum?id=r1Ddp1-Rb} {mixup: Beyond
  empirical risk minimization}.
\newblock In \emph{International Conference on Learning Representations}.

\bibitem[{Zhang et~al.(2017)Zhang, Rudinger, Duh, and
  Durme}]{Zhang2017OrdinalCI}
Sheng Zhang, Rachel Rudinger, Kevin Duh, and Benjamin~Van Durme. 2017.
\newblock Ordinal common-sense inference.
\newblock \emph{Transactions of the Association for Computational Linguistics},
  5:379--395.

\bibitem[{Zhang et~al.(2021{\natexlab{a}})Zhang, Gong, and
  Choi}]{zhang2021capturing}
Shujian Zhang, Chengyue Gong, and Eunsol Choi. 2021{\natexlab{a}}.
\newblock Capturing label distribution: A case study in nli.
\newblock \emph{arXiv preprint arXiv:2102.06859}.

\bibitem[{Zhang et~al.(2021{\natexlab{b}})Zhang, Gong, and
  Choi}]{zhang-etal-2021-knowing}
Shujian Zhang, Chengyue Gong, and Eunsol Choi. 2021{\natexlab{b}}.
\newblock \href {https://doi.org/10.18653/v1/2021.findings-acl.172} {Knowing
  more about questions can help: Improving calibration in question answering}.
\newblock In \emph{Findings of the Association for Computational Linguistics:
  ACL-IJCNLP 2021}, pages 1958--1970, Online. Association for Computational
  Linguistics.

\bibitem[{Zhou et~al.(2004)Zhou, Bousquet, Lal, Weston, and
  Sch{\"o}lkopf}]{zhou2004learning}
Dengyong Zhou, Olivier Bousquet, Thomas~N Lal, Jason Weston, and Bernhard
  Sch{\"o}lkopf. 2004.
\newblock Learning with local and global consistency.
\newblock In \emph{Advances in neural information processing systems}, pages
  321--328.

\bibitem[{Zhou et~al.(2021)Zhou, Nie, and Bansal}]{zhou2021distributed}
Xiang Zhou, Yixin Nie, and Mohit Bansal. 2021.
\newblock Distributed nli: Learning to predict human opinion distributions for
  language reasoning.
\newblock \emph{arXiv preprint arXiv:2104.08676}.

\bibitem[{Zhu et~al.(2019)Zhu, Ko, and Mak}]{zhu2019mixup}
Yingke Zhu, Tom Ko, and Brian Mak. 2019.
\newblock Mixup learning strategies for text-independent speaker verification.
\newblock In \emph{Interspeech}, pages 4345--4349.

\end{thebibliography}
\bibliographystyle{acl_natbib}

\clearpage
\onecolumn
\section*{Appendix}

\appendix\section{Hyperparameters and Experimental Settings}
\subsection*{NLI Hyperparameters and Experimental Settings}
Our implementation is based on the \textit{HuggingFace Transformers}~\citep{wolf2020transformers}.
We optimize the KL divergence as objective with the Adam optimizer \citep{kingma2014adam} and batch size is set to 128 for all experiments.
The Roberta-base is trained for $3,500$ iterations on single-annotated data.
For the finetuning phase, the model is trained for another $30$ iterations. 
The learning rate, $10^{-5}$,  is chosen from AllenTune \citep{dodge2019show}.
For MixUp, the number of training iteration is $3,500$. The $\eta$ of the Beta$(\eta, \eta)$ distribution is $1$.
We choose the same batch size 128 for single label, multi label, and unlabeled data. Thus it will generate evenly interpolated examples. We set the maximum value of loss weight $\alpha$ as $2.0$ and linearly ramp up $\alpha$ from $0$ to its maximum value over the first $100$ iterations of training as is common practice \cite{tarvainen2017mean, berthelot2019mixmatch}.


\subsection*{UFET Hyperparameters and Experimental Settings}
Following the settings from \citet{choi2018ultra}, we set the LSTMs' dimension as $100$. For word vectors, we use $300$ dimensional pretrained Glove. For location vectors, we use 50 dimensions. For sentence length, we cut off the sentence after 50 tokens. For mentions spans, we cut off after 25 characters and ignore mentions longer than 10 words during training. 
Dropout is use for regularization with a probability of $0.5$ for mention representations and $0.2$ for input sentences. We set the batch size as $1000$. Adam optimizer \cite{kingma2014adam} is utilized for optimizing the model parameter with initial learning rate of $0.001$. For MixUp, we follow the same settings in the NLI experiments. The number of training iteration is $10,000$. The $\eta$ of the Beta$(\eta, \eta)$ distribution is $1$. Same batch sizes are chosen for single label, multi label, and unlabeled data. The maximum value of loss weight $\alpha$ is set as $2.0$.

\section{Full Experimental Results}
\begin{table*}[h]
\centering
\scriptsize
    \centering
   \resizebox{1.0\columnwidth}{!}{\begin{tabular}{l|l|r|r|r|r|r|r|r|r|r}
\toprule
& & \multicolumn{9}{c}{Number of Total Annotations}\\
 &  & \multicolumn{3}{c|}{ 150k} & \multicolumn{3}{c|}{ 15k}  &   \multicolumn{3}{c}{ 6k} \\ 
  Data & Learning & JSD$\downarrow$  & KL $\downarrow$ &  acc (old/new)$\uparrow$ & JSD $\downarrow$ & KL $\downarrow$ &  acc (old/new) $\uparrow$   & JSD$\downarrow$  & KL $\downarrow$ &  acc (old/new)$\uparrow$ \\ \midrule

\single & CE& 0.252	&0.548&	0.670 / 0.670& 0.264 & 0.569 & 0.648 / 0.650 & 0.283	&0.556&	0.632 / 0.626 \\ 
\single & \mxone{s} & 0.251 & 0.470 & 0.672 / 0.682 & 0.263 & 0.566 & 0.646 / 0.654 & 0.277 & 0.544 & 0.628 / 0.626 \\ 
\single + \multi & CE (combined) & 0.240 & 0.355 & 0.676 / 0.672 & 0.268 & 0.438 & 0.642 / 0.654 & 0.279 &  0.502 & 0.633 / 0.628 \\
\single + \multi & CE (upsampling) & 0.245 & 0.292 & 0.664 / 0.674 & 0.261 & 0.371 & 0.620 / 0.660 & 0.270 & 0.491 & 0.618 / 0.620  \\
\single + \multi & CE (\single then \multi) &  \textbf{0.217} & \textbf{0.227} & 0.685 / \textbf{0.722} & 0.254 & \textbf{0.285} & 0.628 / \textbf{0.668} & 0.272 & 0.496 & {0.636} / 0.629 \\

\single + \multi &\mxtwo{s}{m} &0.233& 0.285 &	0.682 / 0.682& 0.252 & 0.384 & 0.662 / 0.658 &0.267& 0.490 &	0.610 / 0.636\\ \midrule
\single + \unlabel & \mxtwo{s}{u} & 0.251 &0.472&	0.672 / 0.670& 0.264 & 0.492 & 0.660 / 0.656 & 0.275 &0.504&	\textbf{0.638} / 0.628\\
\single + \multi + \unlabel &\mxtwo{s}{u} then \multi & 0.250 & 0.454 & 0.674 / 0.674 & 0.263 & 0.461 & 0.662 / 0.660 & 0.270 & 0.496 &  0.632 / 0.636   \\
\single + \multi + \unlabel&\mxthree{s}{m}{u} & 0.232& 0.283 & \textbf{0.686} / 0.694 & \textbf{0.248} & 0.341 & \textbf{0.668} / 0.666 & \textbf{0.266}& \textbf{0.392} & 0.602 / \textbf{0.642} \\
 \bottomrule
    \end{tabular}}
    \caption{Performance on the {\bf ChaosSNLI} dataset development set. Each column block (150k, 15k, 6k) shows the number of total training annotations. All results use the same amount of annotations, and each row block uses roughly same amount of training examples (bottom row block incorporates large unlabeled data).  }
    \label{tab:snli_fullresult}
\end{table*}

\begin{table}[h]
\scriptsize
    \resizebox{1.0\columnwidth}{!}{\begin{tabular}{l|l|r|r|r|r|r|r|r|r|r}
\toprule
& & \multicolumn{9}{c}{Number of Total Annotations}\\
 &   & \multicolumn{3}{c|}{150k} & \multicolumn{3}{c|}{15k} & \multicolumn{3}{c}{6k} \\ 
  Data & Learning   & JSD$\downarrow$  & KL $\downarrow$ &  acc (old/new)$\uparrow$ & JSD $\downarrow$ & KL $\downarrow$ &  acc (old/new) $\uparrow$   & JSD$\downarrow$  & KL $\downarrow$ &  acc (old/new)$\uparrow$ \\ \midrule
  
\single & CE&  0.312	&0.572&	0.628 / 0.578& 0.319 & 0.686 & 0.552 / 0.528 & 0.330 & 0.753 & 0.516 / 0.526\\ 
\single & \mxone{s} & 0.300 & 0.567 & 0.628 / 0.580 & 0.315 & 0.694 & 0.555 / 0.530 & 0.321 & 0.696 & 0.518 / 0.528\\ 
\single +  \multi &CE(combined)  & 0.256 & 0.370 & 0.626 / 0.584 & 0.269 & 0.393 & 0.550 / 0.530 & 0.302 & 0.422 & 0.520 / 0.532\\
 \single + \multi & CE(upsampling) & 0.249 & 0.293 & 0.614 / 0.610 & 0.251 & 0.341 & 0.545 / \textbf{0.588} & 0.285 & 0.421 & 0.506 / 0.528 \\
 \single + \multi &CE(\single then \multi) &  \textbf{0.213} & \textbf{0.216} & \textbf{0.638 / 0.646} & \textbf{0.246} & \textbf{0.258} & \textbf{0.560} / 0.562 & 0.298 & 0.414 & 0.519 / 0.531\\

 \single + \multi & \mxtwo{s}{m} &0.243& 0.288 &	0.598 / 0.602& 0.254 & 0.357 & 0.534 / 0.568 & 0.271 & 0.409 & 0.520 / 0.539\\ \midrule
\single + \unlabel & \mxtwo{s}{u} & 0.294 &0.537&	0.626 / 0.566& 0.301 & 0.539 & 0.544 / 0.560 & 0.309 & 0.617 & 0.519 / 0.529 \\

\single + \multi + \unlabel &\mxtwo{s}{u} then \multi & 0.290 & 0.510 & 0.626 / 0.570 & 0.290 & 0.491 & 0.554 / 0.564 & 0.295 & 0.571 & 0.521 / 0.533 \\
\single + \multi + \unlabel&\mxthree{s}{m}{u}  &  0.241& 0.287 & 0.596 / 0.610 & 0.252 & 0.348 & 0.548 / 0.570 & \textbf{0.266} & \textbf{0.384} & \textbf{0.522 / 0.540}\\
 \bottomrule
    \end{tabular}}
    \caption{Performance on the {\bf ChaosMNLI} dataset development set. Each column block (150k, 15k, 6k) shows the number of total training annotations. All results use the same amount of annotations, and each row block uses roughly same amount of training examples (bottom row block incorporates large unlabeled data). }
    \label{tab:mnli_fullresult}
\end{table}



\section{Label Count Comparison}
\begin{table}[h]
\centering
    \centering
    \begin{tabular}{l|l|r|r|r|r}
\toprule
\# Multi  & \# Single &JSD & KL &  acc (old/new) & $H$ \\ \midrule
0 &150K& 0.25&0.55&	0.676 / 0.688&	0.363 \\
  0.5K (20-way)  & 130K & 0.20&0.22&	0.676 / 0.726&	0.695\\
  1K (10-way) &140K& 0.19&0.22&	0.684 / 0.732&	0.643\\ 
  5K (5-way) &145K& 0.19&0.22&	0.676 / 0.732&	0.701 \\ 
 \bottomrule
    \end{tabular}
    \caption{Label count comparison on ChaosSNLI dataset. The total number of labels is consistent among different rows (150K). $H$ represents the predicted label entropy.}
    \label{tab:label_count_comparison}
\end{table}

\section{Training Data Configuration for 6K NLI}
\begin{table}[h]
\centering
\footnotesize
 \begin{tabular}{l|l|c|c|c|c|c}
 \hline
 {Task} & {Data Setup}&   \# {Single} & \# {Multi}  & \# {Unlabel}  & {Total \# Labels} & {Total \# Examples}\\ \hline
 & Original & 549k / 392k & 0& 0 & 549k / 392k& 549k / 392k  \\\cline{2-7}
 & \single &6k & 0 & 0 &  6k * 1 = 6k& 6k\\
Chaos  & \single + \multi &  1k & 0.5k &  0& 1k * 1 + 0.5k * 10  = 6k & 1.5k \\
S / MNLI& \single + \unlabel & 6k  &0 & 549k-6k & 6k * 1  = 6k & 549k \\
  &  \single + \multi + \unlabel 
 & 1k & 0.5k & 549k-1.5k&  1k * 1 + 0.5k * 10  = 6k& 549k \\
 \hline
  \end{tabular}\vspace{-0.3em}
 \caption{Training data configurations for 6k NLI. Each configuration is characterized by the number of labels and the number of examples. The number of labels are consistent in all settings. In NLI task, each multi label example contains 10 labels. For completeness, we also provide original training data configurations.}  \vspace{-0.3em}
    \label{tab:data_setting_6k}
\vspace{-3mm}    
\end{table}

\end{document}